\documentclass[reprint,
amsmath,amssymb,
aps,
prl, 
floatfix,
footinbib,
]{revtex4}

\usepackage{booktabs,siunitx}
\usepackage[usenames,dvipsnames]{xcolor}
\usepackage{graphicx}
\usepackage{dcolumn}
\usepackage{bm}
\usepackage{float}
\usepackage{natbib}
\usepackage{hyperref}
\usepackage{xr}
\usepackage{cleveref}

\usepackage{wrapfig}
\usepackage{comment}
\usepackage[protrusion=false,expansion=true,kerning=true,babel=true,final]{microtype}
\usepackage{soul}
\usepackage{lettrine}
\usepackage{lipsum} 

\graphicspath{{Figures/}}

\usepackage{tikz}
\usetikzlibrary{matrix}

\newcommand{\fix}[1]{\noindent  {\colorbox{BrickRed}{\color{White} 
FIX:}  \color{BrickRed}#1\normalcolor}}

\begin{document}
\setlength{\belowcaptionskip}{-3ex plus .5ex minus .5ex}
\setlength{\parskip}{1ex plus .25ex minus .25ex}
\setlength{\abovedisplayskip}{1ex plus .05ex minus .05ex}%
\setlength{\belowdisplayskip}{1ex plus .05ex minus .05ex}%
\setlength{\abovedisplayshortskip}{0pt}%
\setlength{\belowdisplayshortskip}{0pt}


\author{Mrunmayi Mungekar$^{1}$}
\author{Sanjith Menon$^{1}$}
\author{M. Ravi Shankar$^{2}$}
\author{M.K. Jawed$^{1}$}
\affiliation{\footnotesize 
$^1$Dept.\ of Mechanical and Aerospace Engineering, University of California, Los Angeles, CA 90095, USA \\
$^2$Dept.\ of Industrial Engineering, University of Pittsburgh, Pittsburgh, PA, 15261 USA \\
$^*$ Corresponding author: M.R.S.: ravishm@pitt.edu and M.K.J.: khalidjm@seas.ucla.edu}

\begin{abstract}
%
We present a simple, accessible method for autonomously transforming flat plastic sheets into intricate three-dimensional structures using only uniform heating and common tools such as household ovens and scissors. Our approach combines heat-shrinkable thermoplastics with Kirigami patterns tailored to the target 3D shape, creating bilayer composites that morph into a wide range of complex structures, e.g., bowls, pyramids, and even custom ergonomic surfaces like mouse covers. Critically, the transformation is driven by a low-information stimulus (uniform heat) yet produces highly intricate shapes through programmed geometric design. The morphing behavior, confirmed by finite element simulations, arises from strain mismatch between the contracting thermoplastic layer and the constraining Kirigami layer. By decoupling material composition from mechanical response, this method avoids detailed process control and enables a broad class of self-morphing structures, offering a versatile platform for adaptive design and scalable manufacturing.
\end{abstract}

\pacs{Valid PACS appear here}

\title{Directed Shape Morphing using Kirigami-enhanced Thermoplastics}

\maketitle

\newcommand{\latentdynamics}{reduced-order dynamics}

\begin{figure}[b!]
    \centering
    \includegraphics[width=0.8\textwidth]{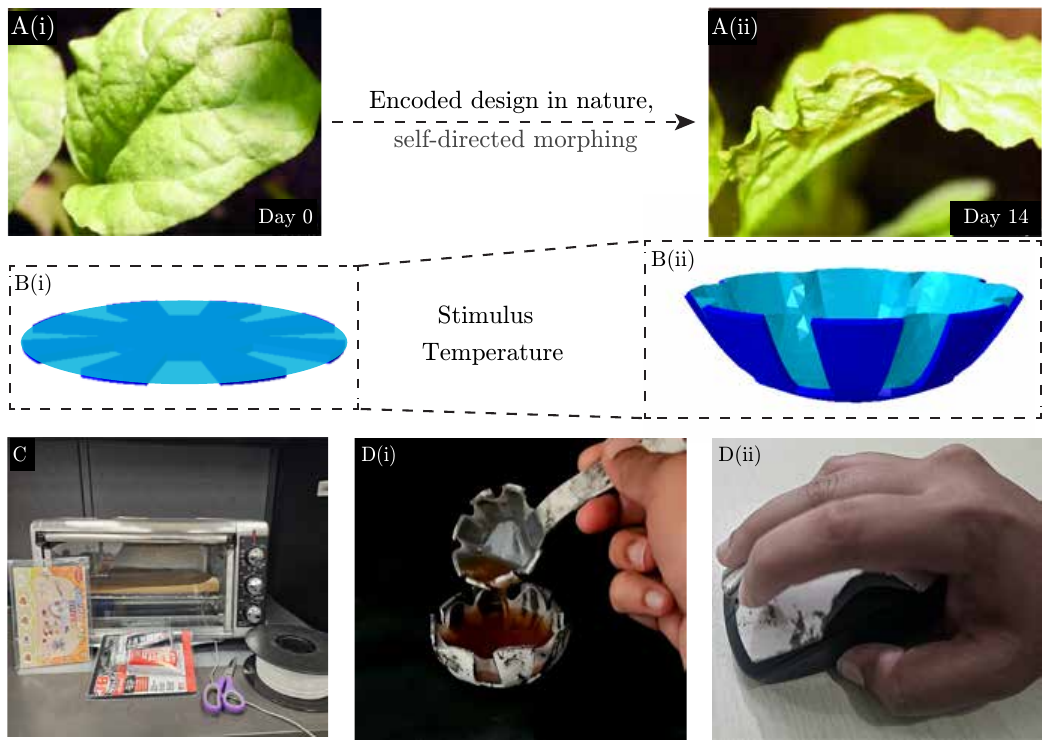}
    \caption{\textbf{Introduction to Kirigami-Enhanced Shrinky Dink 3D Morphing.}
    A(i–ii) Nature exhibits growth-based, interference-free morphogenesis in various organisms (images reproduced with permission from Sharon et al.~\cite{sharon2004leaves}).
    B(i–ii) Shrinky Dink deployables mimic these natural processes, morphing from one form to another without design interference. 
    (C) This technology enables the use of common equipment, such as a domestic oven and glue, to create a wide range of functional structures, including D(i) a soup bowl and spoon, and D(ii) a mouse cover.}
    \label{fig:overview}
\end{figure}

Traditional manufacturing processes define the geometry of engineered components by prescribing the contours of the structure -- be it by defining the mold geometry in processes like casting, forging or extrusion or by controlling the tool path during subtractive processes (e.g., milling, turning or grinding). Additive manufacturing processes define the part volumetrically -- voxel-by-voxel in methods that range of stereolithography, direct ink writing or laser/electron-beam powered powder sintering \cite{kalpakjian2009manufacturing}. 
Natural systems have mastered subtlety in programming shape selection. 
Structural evolution is a composite response resulting from interplay of growth, boundary conditions and mechanical anisotropy \cite{oliver2016morphing}. Shape selection is emergent rather than one arising from information dense path-planning. 
Morphogenesis guided by genetic, mechanical  and environmental cues shapes organisms across length scales in both animals and plants. They impact the supercoiling of DNA plasmids \cite{thompson2002supercoiling}, embryonic development \cite{hogan1999morphogenesis, Baguma-Nibasheka2023-fa}, define the intricacy of the cerebral cortex \cite{Reyssat_Mahadevan_2009, Rebocho2017}, direct the coiling of tendrils \cite{gerbode2012cucumber}, open pine cones \cite{eger2022structural}, seed-pods \cite{armon2011geometry} and even hierarchically ruffle leaves via anisotropic growth \cite{ sharon2004leaves} (Fig.~\ref{fig:overview}A). The heterogeneity of the growth along the edges transforms a leaf from a hitherto flat profile (Fig.~\ref{fig:overview}A(i)) into one decorated along its periphery by wave-like, buckled features (Fig.~\ref{fig:overview}A(ii)).

Where form defines functionality, structural morphing enables adaptivity in engineered systems. Fabricating structures from active and responsive materials has enabled applications spanning  aerospace~\cite{space1furuya1992concept,space2hanaor2001evaluation,space3santiago2013advances}, medicine \cite{med1du2018programmable,med2banerjee2020origami, babaee2019temperature}, and intelligent machines\cite{im1wang2017modular,im2chi2022bistable} assimilate responsive materials to tune their functionality. Motivated by natural systems, the mechanical responsiveness in engineered structures has been explored as a pathway for shape-selection. Mechanical boundary conditions and buckling have emerged as a versatile tool for deriving 3D shapes by patterning bending strains~\cite{park2021three, mouthuy2012overcurvature}. In materials such as liquid crystalline elastomers (LCE), the ability to pattern the microstructure has allowed for stimuli-driven triggering of metrics that are incompatible in a hitherto flat sheet. 3D shapes emerge from 2D monoliths with a structural resolution that is only limited by the scale of the blueprinted microstructure~\cite{aharoni2018universal, ware2015voxelated, kowalski2017voxel}. Mechanical nonlinearities resulting from anisotropic material response has the potential to subvert the resolution limit placed by the material's microstructure. A remarkable example is the demonstration of hierarchically buckled and wrinkled structures that assemble in gels due to (radial) shrinkage gradients that are azimuthally symmetric~\cite{klein2007shaping}. Complexity emerges at scales that are finer than that dictated by the microstructural inhomogeneity  due to the interplay of stimulus response and mechanical non-linearity. This has also been observed in topologically confined LCE~\cite{clement2021complexity}, assemblies of segmented rings~\cite{lu2024multistability} and growth of filaments in confined volumes~\cite{vetter2014morphogenesis}. These observations harken back to phenomena observed in the natural world, where strain generation in the presence of topological/geometric confinement drives morphogenesis that is emergent from mechanical non-linearity. 

The transformation of flat preforms into 3D shapes has relied on the competition between stretching and bending. A trade-off emerges in slender objects between stretch whose energy scales with thickness, versus that for bending out of the plane, whose energy penalty scales with thickness-cubed~\cite{armon2011geometry}. Inducement of strains that are incompatible in 2D can be used to generate 3D shapes -- a theme that recurs in multiple examples~\cite{aharoni2018universal,pyramid,ware2015voxelated,modes2011gaussian,pezzulla2016geometry}. The ability to pattern the distribution of bending and stretching strains in a flat structure expands the design space. It becomes possible to explicitly design curvatures, when needed by dictating the gradient in strain through the thickness and rely on the patterning of stretching strains to assemble 3D shapes, when desired. Consider, the transformation of a flat-shape (Fig.~\ref{fig:overview}B(i)) into a bowl (Fig.~\ref{fig:overview}B(ii)) when subjected to say, a temperature stimulus. Intuitively, it is evident that the transformation of a flat structure into a 3D shape requires generation of thermomechanical strains that are incompatible in the hitherto 2D preform. But, the generation of the curved sides of the bowl requires the ability to encode the corresponding bending strains. While, this represents a simple example, it motivates the programming of structural hierarchies emerging from spatially-indexed patterning of bending of stretch in flat preforms. 

Here, we leverage a bilayer approach that patterns bending and stretching strains by compositing shape-memory thermoplastics with a passive Kirigami layer, enabling precise spatial control over thermomechanical shape selection from flat freeforms. Rather than tailoring material microstructure, our method decouples material response from mechanical design by using a homogeneous thermoplastic substrate and programming deformation solely through Kirigami patterning. Building upon this, we employ a physically informed machine-learning framework~\cite{inversedesign} to assemble arbitrary 3D shapes via optimized Kirigami designs. Our experiments utilize prestrained polystyrene (commonly known as Shrinky Dinks) composited with a Kirigami layer (ABS plastic) through gluing. The Kirigami defines the local strain distribution under uniform heat treatment: regions unconstrained by Kirigami undergo shrinkage (stretch), while constrained areas generate bending strains. Patterning these one-way shape memory strains within the bilayer directs programmable shape selection. This approach generalizes to a broad range of thermally triggered, one-way shape memory polymers, including polyolefins~\cite{nguyen2010better}, polylactic acid and its co-polymers~\cite{yahia2015shape}, biodegradable polyester–polyurethane networks~\cite{alteheld2005biodegradable}, perfluorosulphonic acid ionomers~\cite{xie2010tunable}, and polymethylmethacrylate~\cite{wang2019influence}, among others; see Ref.~\cite{liu2007review} for a comprehensive classification. The passive Kirigami layer can be fabricated via manual, laser, or machining methods, printed as a single sheet, or stamped for large-scale manufacturing. Crucially, the entire fabrication process is accessible with household equipment (Fig.~\ref{fig:overview}C). Prototypical shapes such as bowls and spoons (Fig.~\ref{fig:overview}D(i)) and functional ergonomic surfaces like a computer mouse cover (Fig.~\ref{fig:overview}D(ii)) demonstrate the method’s versatility. Kirigami–substrate composites designed through our algorithmic framework enable compact, deployable structures ideal for applications requiring volume-efficient storage and on-demand deployment—such as disaster relief or space travel -- while inherently supporting scale-free design from micrometer to meter scales.

\noindent \textbf{\large Fabricating and Deploying Kirigami-Composites}

\noindent \textbf{Methodology:} 
In this study, we harness the thermal responsiveness of Shrinky Dinks -- prestrained polystyrene sheets that contract laterally and thicken when heated, shrinking to roughly one-third their original size -- to induce self-directed structural morphing. Departing from traditional uses that apply inks or 2D prints directly on the Shrinky Dink surface, we form a bilayer composite by adhering a separately fabricated Kirigami layer onto the substrate. It is important to emphasize that the Kirigami layer itself is passive and does not undergo any intrinsic shape change upon heating. Unlike 4D printing approaches, which rely on programmed material deformation triggered during fabrication or thermal stimulus, our method uses 3D printing solely as a fabrication tool for the Kirigami pattern, which mechanically constrains the Shrinky Dink’s contraction without altering its own shape. This clear distinction underscores that the shape transformation results from the interplay between the thermally shrinking Shrinky Dink and the passive Kirigami constraints.

Upon uniform heating, the Shrinky Dink attempts to contract, but the attached Kirigami layer locally restricts shrinkage, inducing stress-driven buckling that transforms the initially flat composite into complex 3D shapes. By strategically designing the Kirigami patterns, we control where contraction is permitted and where bending strains develop, enabling precise, programmable formation of diverse 3D structures. To fabricate these Kirigami patterns and rapidly explore design variations, we employ 3D printing; the resulting shape change arises from the mechanical interaction between the Shrinky Dink's contraction and the Kirigami layer, which bends and deforms but does not undergo any intrinsic shape change on heating. The following subsection details the experimental process by which these principles are implemented. The following subsection details the experimental process by which these principles are implemented.

\begin{figure}[b!]
    \centering
    \includegraphics[width=0.8\textwidth]{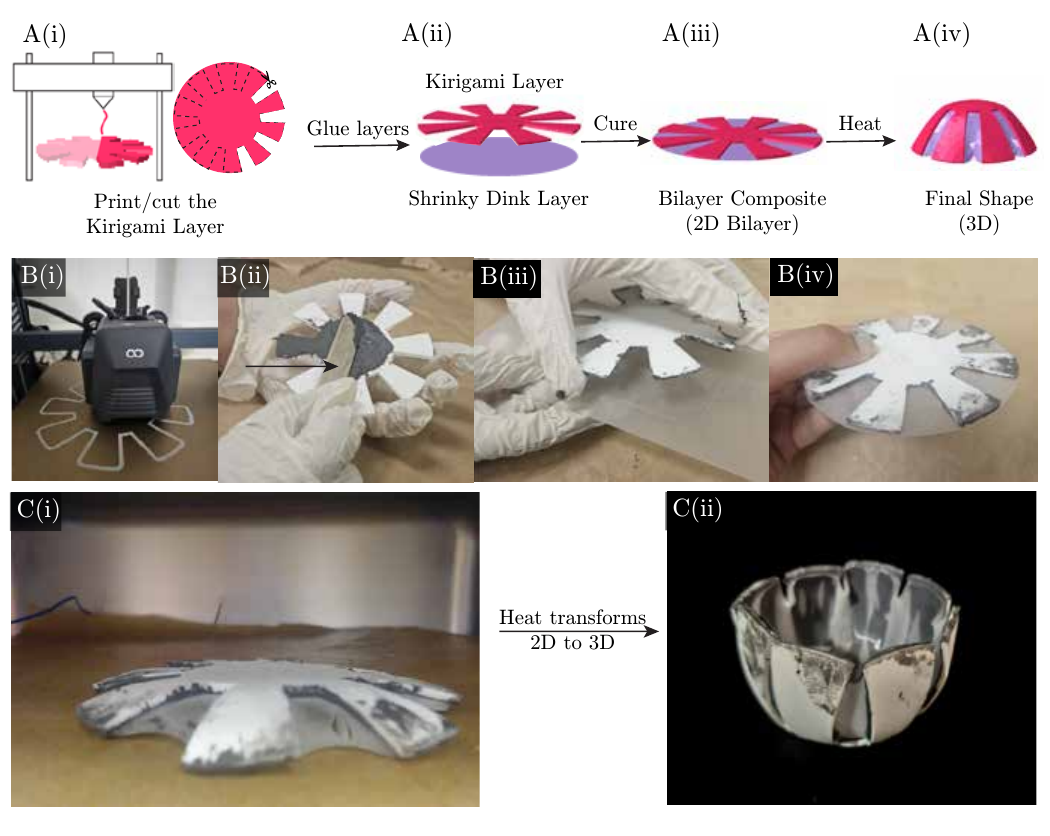}
    \caption{\textbf{Methodology for manufacturing Shrinky Dink deployables.} 
    (A) Schematic of the process: starting with the Kirigami design, gluing the two layers together, and finally heating the composite to achieve the 3D structure.
    (B) Experimental images illustrating each step: B(i) 3D-printed Kirigami, B(ii) application of glue to the Kirigami layer, B(iii) bonding the two layers, and B(iv) the final cured Shrinky composite.
    (C) Deployment of the structure: C(i) buckling of the composite in the oven, and C(ii) the fully formed 3D shape after deformation.}
    \label{fig:experiment}
\end{figure}

\vspace{2ex}
\noindent \textbf{Experimental Manufacturing:}
The manufacturing process begins with the 3D printing of a carefully selected Kirigami design (Fig.~\ref{fig:experiment}(B(i))) using ABS filament, chosen for its durability and heat resistance. After printing, one side of the Kirigami layer is gently roughened with sandpaper or a hand file to enhance adhesion. In parallel, a piece of Shrinky Dink sheet is cut into the required shape -- typically circular to match the final 3D structure -- and similarly roughened on one side. Prior to bonding, both surfaces are thoroughly cleaned with 99\% isopropyl alcohol to ensure effective adhesion. Bonding is performed by first applying a thin layer of epoxy glue to the roughened surface of the 3D printed Kirigami piece. The glued side is then carefully aligned with and pressed onto the Shrinky Dink substrate (Fig.~\ref{fig:experiment}(B(ii))). The assembly is subjected to curing under a substantial load for at least 24 hours to secure a robust bond. After curing, the planar composite can be stored virtually indefinitely and deployed as needed.

To deploy the composite (Fig.~\ref{fig:experiment}(B(iii))), the sample is heated in a domestic oven preheated to 270$^\circ$F -- deliberately lower than the 300–325$^\circ$F typically used for Shrinky Dink toys -- to slow the contraction process and afford greater control over the buckling phenomenon. The composite is placed with the Shrinky Dink side facing down to promote controlled deformation. Despite the reduced temperature, the transformation into the final three-dimensional shape generally occurs within 2–3 minutes.

This approach relies on cost-effective, widely accessible tools and materials: a basic home 3D printer for Kirigami fabrication, commercially available Shrinky Dink sheets, strong epoxy adhesives, and a standard household oven. Supporting items such as isopropyl alcohol, sandpaper, and scissors are inexpensive and readily available. However, using domestic equipment introduces experimental variability due to imprecise temperature control, while the commercial nature of materials like Shrinky Dinks and ABS—with their complex and sometimes variable compositions—can further affect the consistency and predictability of results.. We provide a detailed account of experimental sources of error and modeling simplifications in the supplementary material, including temperature fluctuation data (Supplementary Material S1) and material modeling details (Supplementary Material S2). The epoxy occasionally appears smudged and aesthetically imperfect, a consequence of prioritizing procedural simplicity; these surfaces can be painted post-fabrication to restore appearance without affecting the final structure.

\vspace{2ex}
\noindent \textbf{\large Numerical Characterization}\\
To complement our experimental observations, we developed a finite element simulation framework to capture the morphing behavior of the bilayer composites. Accurate material modeling poses two main challenges. First, off-market Shrinky Dink sheets exhibit significant variability in material properties, with nonlinear and orthotropic stress–strain characteristics that are difficult to characterize. Second, the properties of the 3D-printed Kirigami layer slightly depend on fabrication parameters such as printing temperature, infill rate, and pattern, leading to additional heterogeneity. These complexities can result in inhomogeneous thermal contraction and loss of symmetry in the final structures. To make the problem tractable, we introduce a few simplifying assumptions in the simulation framework to address these issues.

Material modeling details, including parameter values, are provided in the Supplementary section. We modeled the Shrinky Dink layer with a Young's modulus of $E^s = 404$~MPa, thermal contraction coefficient $\alpha^s = -0.01$~K$^{-1}$, and thickness $t^s = 0.1$~mm. For the Kirigami layer, we used $E^k = 762$~MPa and $t^k = 1.8$~mm, with thermal expansion neglected ($\alpha^k = 0$~K$^{-1}$), as the measured value ($0.0001$~K$^{-1}$) is insignificant relative to the Shrinky Dink\rq{}s thermal response. In all cases, we assume linear elastic behavior and neglect any temperature-dependent variation in Young's modulus, even though both layers may exhibit mild nonlinearity and temperature dependence in practice.

For mesh generation and model implementation, we constructed each composite as two solid material layers composed of 6-node linear triangular prism (C3D6) elements, with shared nodes at the interface to simulate perfect adhesion. Planar meshes are first generated in MATLAB and then stacked to form 3D solid elements. Simple shapes were defined directly using MATLAB's geometric functions, whereas complex Kirigami patterns were created in CAD software, exported as STL files, and subsequently imported for mesh generation. All finite element simulations were performed using the commercial package ABAQUS/Standard (SIMULIA).

To model the thermal mismatch between the two layers, we imposed a predefined, biaxial in-plane contraction stress in the Shrinky Dink layer to represent its negative thermal expansion coefficient, while the Kirigami layer was assigned zero in-plane stress due to its negligible thermal response. This approach directly mimics the difference in thermal responsiveness ($\alpha^s$ vs. $\alpha^k$) between the layers. The resulting stress mismatch drives the composite to deform out of plane as it seeks to minimize total strain energy, consistent with classical mechanics of mismatched bilayers. In our finite element implementation, we calculated the contraction stress in the Shrinky Dink layer as
\begin{equation}
    \sigma_s = -\alpha_s E_s \Delta T,
\end{equation}
where $\Delta T = 110$~K is the imposed temperature change used in both experiments and simulations. Both in-plane principal stresses in the Shrinky Dink layer were set to $\sigma_s$. The resulting stress mismatch drives the bilayer to adopt a three-dimensional equilibrium shape. Boundary conditions match the experiment: the out-of-plane displacement of boundary nodes on the Shrinky Dink layer is fixed, with the interior free to deform.  To account for model simplifications, such as neglecting the glue layer, we treated $\alpha_s$ as a fitting parameter and used a slightly higher value ($\alpha^s = -0.01$~K$^{-1}$) in simulation than measured experimentally ($-0.005$~K$^{-1}$).

\vspace{2ex}
\noindent \textbf{\large Programmable 3D Shape Design} \\
By tailoring the geometries of both the Kirigami layer and the Shrinky Dink sheet, we can programmably control the final 3D shape of the composite. While a complete, systematic solution to the inverse design problem remains a challenge due to material and modeling complexities, we demonstrate three complementary strategies for achieving a wide range of 3D shapes: (A) heuristic, intuition-driven design, (B) machine learning-assisted inverse design, and (C) combinatorial assembly of basic shapes. 


\begin{figure}[b!]
    \centering
    \includegraphics[width=0.7\textwidth]{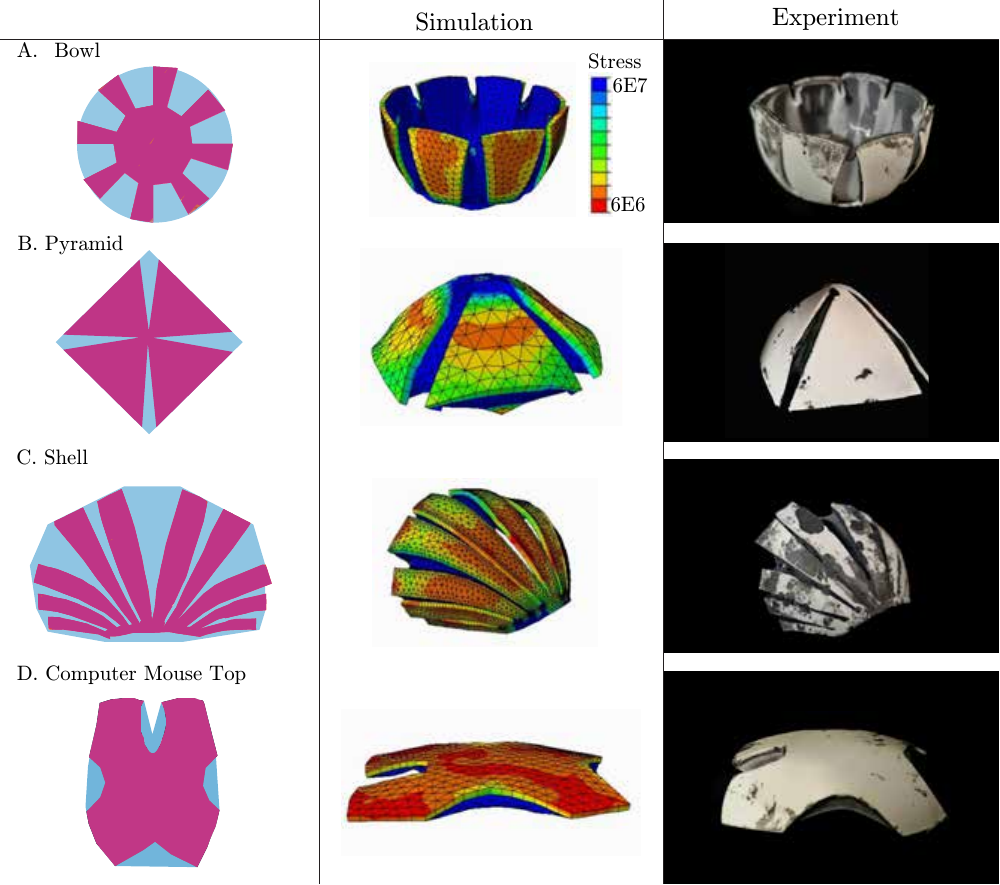}
    \caption{\textbf{Compilation of shapes produced via heuristic, intuition-driven design.} The first column shows the Kirigami and Shrinky Dink patterns used for each structure. The second and third columns display the corresponding numerically simulated and experimentally obtained 3D shapes. The following shapes are demonstrated: A. Bowl, B. Pyramid, C. Shell, and D. Computer mouse top cover.}
    \label{fig:heuristicshapes}
\end{figure}

\vspace{2ex}
\noindent \textbf{A. Heuristic, Intuition-driven Design}\\
We apply heuristic, geometry-based design principles to tailor Kirigami patterns for specific 3D shapes. For example, a lotus-inspired Kirigami with eight equally spaced petals yields a hemispherical bowl in Fig.~\ref{fig:heuristicshapes}(A), as shown in both numerical simulation and experiment. This design provides a foundation for systematically exploring how geometric parameters affect the resulting 3D morphology, as described in the next section. A pyramid is achieved using a straightforward pattern in the Kirigami layer as shown in Fig.~\ref{fig:heuristicshapes}(B). A custom pattern inspired by seashells is used to produce a 3D shell structure in Fig.~\ref{fig:heuristicshapes}(C).


This intuition-driven approach can also be used to fabricate functional objects. For example, to create a computer mouse top cover (Fig.~\ref{fig:heuristicshapes}(D)), we design the Kirigami pattern so that, when combined with the Shrinky Dink and heated, the composite bends along both its length and width. By adjusting the pattern, we control the amount of curvature in each direction to match the domed shape of a mouse. The design includes a cut between the left and right mouse button regions in the composite, so that after morphing, the cover forms two distinct buttons that can be pressed separately, just like a standard computer mouse.


\vspace{2ex}
\noindent \textbf{B. Machine Learning-assisted Inverse Design}\\
Existing design methods that leverage machine learning (ML), such as those developed for soft kirigami deployables~\cite{inversedesign}, can be adapted to our thermally activated system. In the soft systems studied in the literature~\cite{inversedesign}, shape morphing is achieved by combining a prestretched, hyperelastic substrate with a patterned kirigami layer, allowing reversible deformation that is well captured by high-fidelity finite element simulations using established constitutive models such as Mooney-Rivlin. In contrast, our approach uses a thin plastic (Shrinky Dink) that undergoes permanent, thermally driven contraction. Despite this fundamental difference, both systems share a core physical principle: local control of stretch and constraint. Regions without Kirigami cuts undergo contraction, while areas beneath the Kirigami pattern remain constrained, effectively acting as regions of zero thermal contraction. This selective contraction creates a tailored distribution of bending strains and drives the formation of 3D shapes. As a result, design strategies from soft, stretchable kirigami systems can be transferred to rigid, thermally activated plastics with suitable adaptation.

\begin{figure}[h!]
    \centering
    \includegraphics[width=0.7\textwidth]{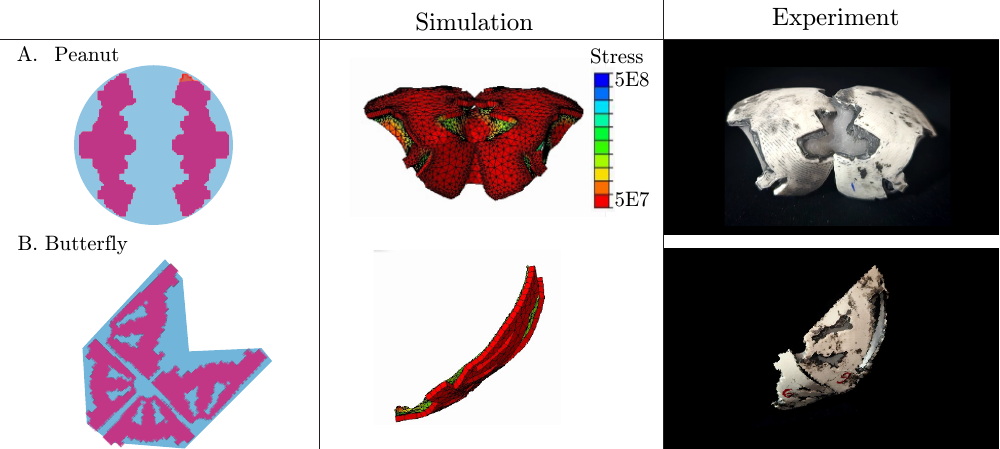}
    \caption{\textbf{Compilation of shapes produced via ML-assisted inverse design.} The first column shows the Kirigami and Shrinky Dink patterns used to produce each shape: A. peanut, and B. butterfly. The second and third columns show the corresponding numerically simulated and experimentally obtained 3D shapes.
    }
    \label{fig:inverseshapes}
\end{figure}

However, directly translating ML-designed Kirigami patterns into experimental shapes presents modeling and manufacturing challenges. Accurate shape prediction may require incorporating additional data -- such as material heterogeneity or process variability -- beyond standard physics-based approaches. Experimentally, inhomogeneity in Shrinky Dink sheets and 3D-printed plastics affects reproducibility and quality control. As a result, adapting these designs requires both geometric tuning and iterative pattern refinement, informed by experimental feedback -- an approach demonstrated using two examples in Fig.~\ref{fig:inverseshapes}A-B.

For the peanut geometry (Fig.~\ref{fig:inverseshapes}A), we directly adopted the Kirigami pattern developed for soft, stretchable composites~\cite{inversedesign}. While this ML-designed pattern yields a peanut shape in the original soft system, translating it to our rigid, thermally activated composite required only geometric adjustment. Specifically, we tuned the overall radius of the Shrinky Dink layer as a fitting parameter to match the experimental outcome. This adjustment is physically motivated: in the soft system, the final 3D shape depends on the ratio of bending and stretching stiffness between the layers, and the optimal Kirigami pattern is sensitive to those ratios. In our system, where material properties differ, scaling the radius serves as a practical proxy for tuning this effective stiffness ratio.

For more complex shapes, such as the butterfly in Fig.\ref{fig:inverseshapes}B, we began with a Kirigami pattern designed to create a flower in soft composites\cite{inversedesign}. To achieve the butterfly form, we applied targeted geometric modifications—such as rotation and aspect ratio adjustment -- guided by design intuition. This demonstrates that, while ML-based inverse design frameworks provide a robust foundation, simple geometric modifications and empirical refinement are often sufficient to realize complex shapes in our rigid, thermally activated composites. Details of the design and fabrication process are provided in the Supplementary Material.

\begin{figure}[h!]
    \centering
    \includegraphics[width=0.7\textwidth]{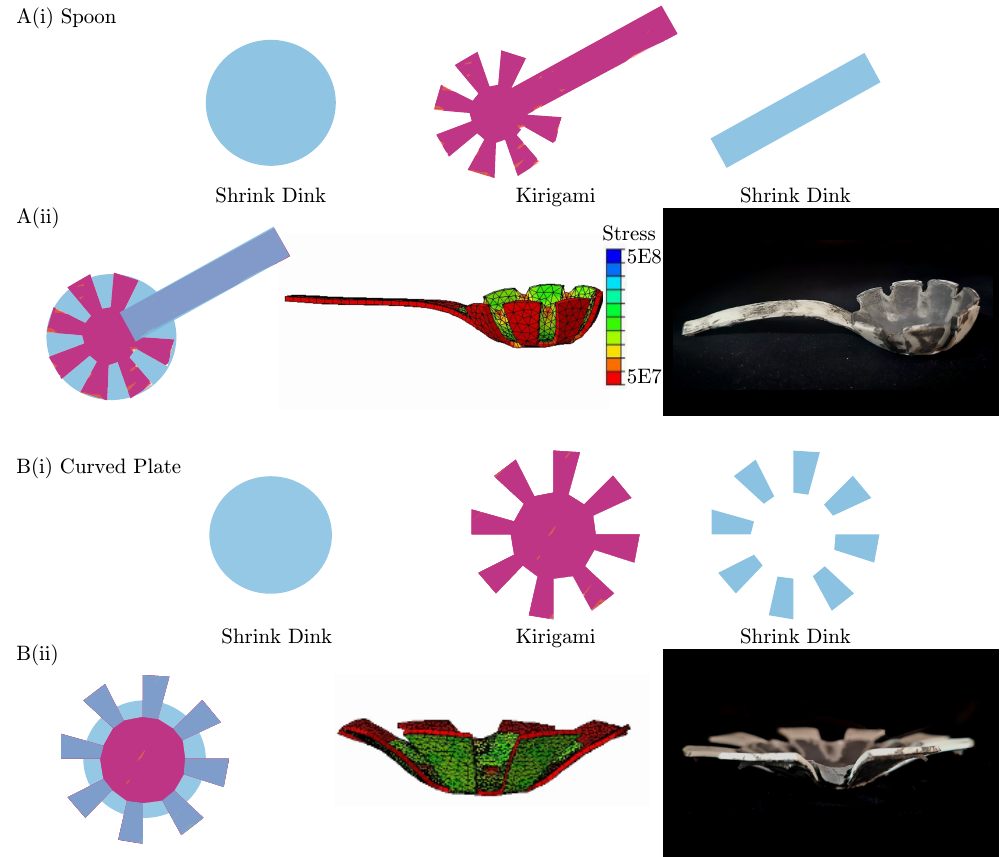}
    \caption{\textbf{Compilation of shapes produced via combinatorial assembly.} The first column illustrates the Kirigami and Shrinky Dink patterns used to fabricate each structure: A. spoon, and B. rimmed plate. A(i) and B(i) show the trilayer construction -- two Shrinky Dink layers and one Kirigami layer -- prior to assembly. In A(ii) and B(ii), the first column shows the undeformed, superposed trilayer composite, while the second and third columns present the numerically simulated and experimentally obtained 3D shapes, respectively.
    }
    \label{fig:combinatorial}
\end{figure}

\vspace{2ex}
\noindent \textbf{C. Combinatorial Assembly of Basic Shapes}\\
The combinatorial assembly approach enables the fabrication of complex 3D structures by joining together basic building blocks -- such as a bowl-like shape and a handle to create a spoon in Fig.~\ref{fig:combinatorial}A -- using multiple Shrinky Dink and Kirigami layers. While we used bilayer composites so far, our method naturally extends to trilayer assemblies and beyond. Fig.~\ref{fig:combinatorial}A shows the process of fabricating a spoon, where we combine two precut Shrinky Dink pieces -- a circular region for the bowl and a rectangular strip for the handle -- with a patterned Kirigami layer sandwiched between the Shrinky Dinks, as shown in Fig.~\ref{fig:combinatorial}A(i). The layers are glued together to form a trilayer composite that, upon heating, transforms into a continuous, functional spoon, as demonstrated by both simulation and experiment in Fig.~\ref{fig:combinatorial}A(ii). A second example, leveraging the same strategy, is the fabrication of a rimmed plate: two Shrinky Dink layers -- one disk for the base and one ring of petals for the rim -- are combined with a patterned Kirigami layer, as shown in Fig.~\ref{fig:combinatorial}B(i). These three layers are assembled and bonded to form a trilayer composite in Fig.~\ref{fig:combinatorial}B(ii). Heating the composite results in a plate with a well-defined raised rim, as seen in both simulation and experimental images.

\begin{figure}[h!]
    \centering
    \includegraphics[width=0.7\textwidth]{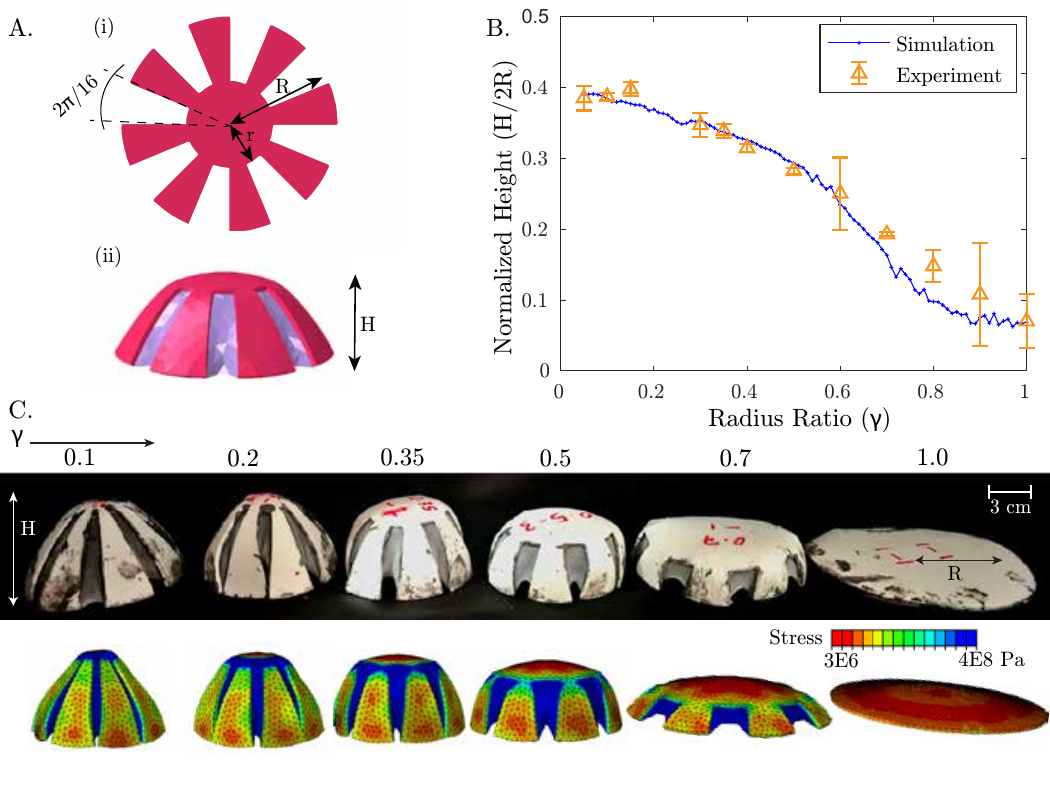}    
    \caption{\textbf{Variation of the 3D shape with change in percentage material removed from kirigami layer }A. A sketch of the various dimensions used in the parametric variation (i) in the planar and (ii) the 3D configurations B. A graph portraying the experimental and the numerical variation of height. C. Variation of height in experimental and numerical specimens with change in $\gamma$}
    \label{fig:parametricvariation}
\end{figure}

\vspace{2ex}
\noindent \textbf{\large Parametric Variation} \\
We demonstrate that the final 3D shape varies systematically with design and manufacturing parameters -- specifically, the Kirigami patterns. Using the previously studied hemispherical bowl shape (Fig.~\ref{fig:heuristicshapes}A) as a baseline, we select the normalized height ($H/2R$) as a quantitative descriptor of the final 3D shape, where $R$ is the radius of the composite in its initial flat configuration and $H$ is the maximum height after morphing (with $H=0$ for the undeformed flat state); see Fig.~\ref{fig:parametricvariation}A. To study this relationship, we fix the outer radius ($R$) and thickness of the bowl and systematically vary the radius ratio $\gamma = r/R,$ where, as shown in Fig.~\ref{fig:parametricvariation}A(i), $r$ is the inner radius of the Kirigami pattern. This parameter $\gamma$ tunes the amount of material removed and, consequently, the degree of constraint on stretching. A smaller $\gamma$ (corresponding to a larger central cut) permits greater contraction and buckling, while a larger $\gamma$ leads to flatter, less deformed structures. Quantitatively, the amount of material removed ($\alpha$) is related to the radius ratio ($\gamma$) by
\begin{equation}
    \alpha = 0.5 \times (1-\gamma^2).
\end{equation}
For each design, we measure the resulting normalized height $H/2R$ in both experiments and simulations. As shown in Fig.~\ref{fig:parametricvariation}B, increasing $\gamma$ systematically decreases $H/2R$, a monotonic trend observed in both experimental and simulated results. Fig.~\ref{fig:parametricvariation}C presents representative experimental and simulated snapshots, showing that as the radius ratio increases (and the amount of material removed decreases), the resulting shape systematically changes and the out-of-plane height decreases.

\vspace{2ex}
\noindent \textbf{\large Conclusions}\\ 
In conclusion, this study presents a versatile and accessible approach for transforming flat thermoplastic sheets into complex 3D via Kirigami patterning and thermal activation. By integrating Shrinky Dink substrates with an inert Kirigami layer, we achieve programmable shape morphing into forms such as bowls, pyramids, and custom objects like mouse covers. This technique relies solely on readily available materials and household tools, enabling precision and scalability without the need for specialized equipment or detailed process control. Our framework establishes a foundation for scalable, self-directed assembly and offers promising opportunities in fields ranging from aerospace and medicine to adaptive and deployable design. Future directions can include further refinement of the design process, with an emphasis on improved material modeling to account for the inherent nonlinearity and heterogeneity of Shrinky Dink substrates, as well as the variability introduced by 3D-printed Kirigami layers. Integrating additional experimental data and real-time process control could enhance the fidelity and reproducibility of shape morphing.


\vspace{1em}
\noindent \textbf{\large References} \vspace{-6em} \\

\vspace{1em}
\noindent \textbf{\large Acknowledgments} \\
We acknowledge financial support from the National Science Foundation (grants: CMMI-2332555 and CMMI-2053971).

\vspace{1em}
\noindent \textbf{\large Author contributions} \\
M.M., M.R.S., and M.K.J. conceived the project and designed the experiments. M.M. and S.M. performed all fabrication and characterization. M.M. and M.K.J. developed and implemented the numerical simulations. M.M. analyzed the data and wrote the first draft. M.R.S. and M.K.J. contributed to revisions. All authors reviewed and approved the manuscript.

\noindent \textbf{\large Competing interests} \\
The authors declare no competing interests.

\end{document}



\maketitle


\section{Software and Data Availability}
The MATLAB codes used to generate meshes and STL files, as well as scripts for running and processing ABAQUS simulations, are available at \url{https://github.com/StructuresComp/Shrinky-Dink}. All data files, including images and datasheets required to reproduce Figs.~3--6 of the main text, are also provided in the repository.

\section{Oven Temperature Logging}
Due to the imprecise nature of the heating of a common, household oven, it is difficult to maintain a constant temperature throughout the experiment. Shrinky Dink sheets also have non-uniform, unpredictable deformation which, coupled with the fluctuating temperature, can lead to varying final results. Recording the temperature fluctuation with respect to time gives anyone who recreates the experiment an idea of how the design parameters fluctuate.

\begin{figure}[h]
    \centering
    \includegraphics[width=\textwidth]{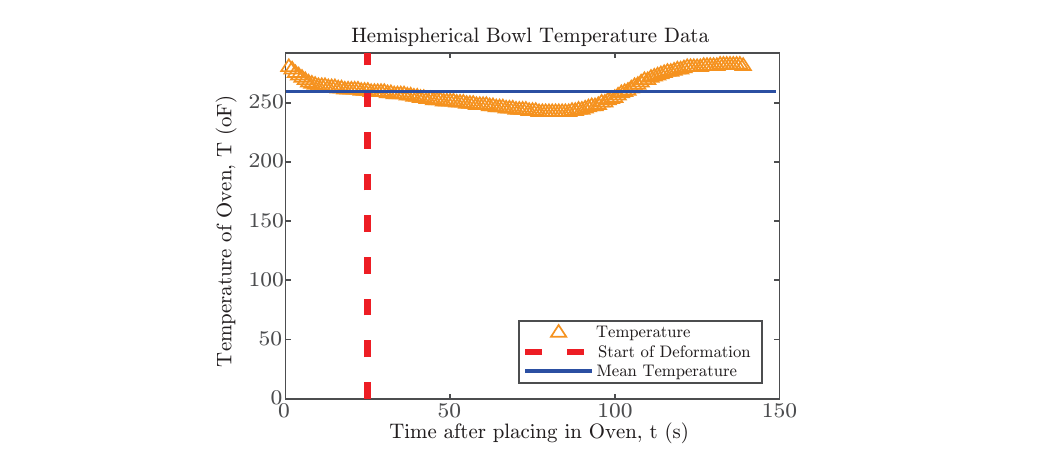}
    \caption{The temperature log data is shown for a hemispherical bowl's deformation. The flucutation about the mean temperature can be seen along with the starting time of the deformation.}
    \label{fig:hemibowl}
\end{figure}

The temperature logging begins as soon as the composite is placed into the oven. The time of initial visible deformation is timed separately to note the beginning of the process. The temperature data is tracked until the composite is removed from the oven. Fig.~\ref{fig:hemibowl} shows the temperature log for a hemispherical bowl.

\section{Material Modelling}
We used Shrinky Dink sheets (Cridoz, purchased from Amazon) and Hatchbox ABS filament for 3D printing the Kirigami layers. All Kirigami samples were printed with 100\% solid infill and a 0.2~mm layer thickness. For material testing, we fabricated dogbone samples with a thickness of 1.8~mm from the Kirigami material.

\begin{figure}[h]
    \centering
    \includegraphics[width=\textwidth]{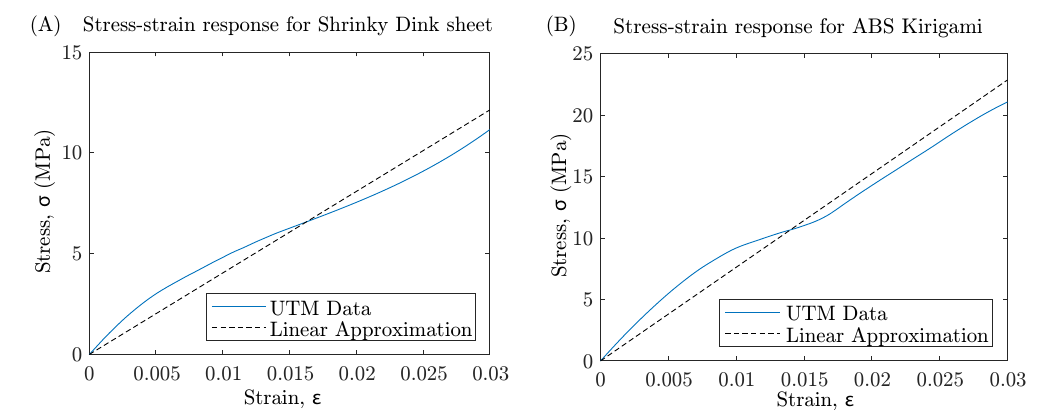}
    \caption{Uniaxial tensile stress–strain responses of (A) Shrinky Dink sheets and (B) ABS Kirigami used in the experiments, as well as the linearized material models.}
    \label{fig:materialmodel}
\end{figure}

We carried out tensile tests on both materials using ASTM D412 dogbone specimens to characterize their stress–strain response. All measured stress–strain curves were nonlinear, as expected. For simulations, we approximated the behavior as linear elastic, fitting the modulus by equating the strain energy of the nonlinear curve and the linear model up to a chosen strain $\epsilon_m$:
\begin{equation}
    \int_{0}^{\epsilon_m} \sigma \,d\epsilon = \frac{1}{2}E\epsilon_m^2
\end{equation}
Fig.~\ref{fig:materialmodel} shows both the measured and the linearized stress--strain curves for the Shrinky Dink and Kirigami layers.

The next phase of material modeling involves determining the thermal expansion coefficients for the two materials. We fabricated 3 cm x 3 cm specimens of both materials and placed them in an oven at 270$^{\circ}$ F. As anticipated, the Shrinky Dink material contracted within its plane and expanded in the perpendicular direction. In contrast, the 3D printed piece, featuring a complex crisscross infill pattern, experienced slight contraction across all three dimensions. We recorded the initial and final lengths, and compute the thermal coefficient based on the temperature difference. This experimental process allowed us to accurately determine the thermal coefficients of the materials.

Table~\ref{tab:materialmodel} summarizes the measured Young’s moduli and thermal expansion coefficients for both materials. In all modeling, we assume the materials are nearly incompressible, implying a Poisson’s ratio of 0.5.

\begin{table}[h]
	\centering
	\begin{tabular}{l c c}
		\midrule
		Parameter  &  Value & Unit \\
		\midrule
		Young's Modulus $E^s$ & 404.2082 & MPa \\
		Thermal Coefficient $\alpha^s$ & -0.005 & K\textsuperscript{-1} \\
		Thickness $t^s$ & 0.1 & mm \\
		\midrule
		Young's Modulus $E^k$ & 761.6368 & MPa \\
		Thermal Coefficient $\alpha^k$ & -0.0001 & K\textsuperscript{-1} \\
		Thickness $t^k$ & 1.8 & mm \\
		\midrule
	\end{tabular}
	\caption{The approximate linear elastic and thermal parameters for the two different materials used: (1) Shrinky Dink (2) ABS Kirigami}
	\label{tab:materialmodel}
\end{table}

\clearpage
\section{Machine Learning-assisted Inverse Design}
\begin{figure}[h]
    \centering
    \includegraphics[width=0.7\textwidth]{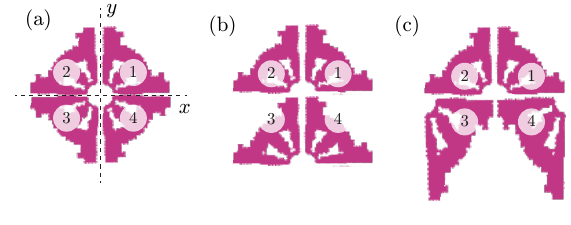}
    \caption{
    \textbf{Heuristic transformation of machine learning–generated Kirigami patterns.}
    (a) The original pattern from Ma et al. (2024) used to generate a flower shape, composed of four identical regions labeled \circled{1}, \circled{2}, \circled{3}, and \circled{4}.
    (b) Heuristic modification by reflecting regions \circled{3} and \circled{4} about the $x$-axis.
    (c) Further adjustment by increasing the aspect ratio of regions \circled{3} and \circled{4}, stretching them along the $y$-axis.}
    \label{fig:SI_MachineLearning}
\end{figure}

As described in the main text (Fig.~4), the butterfly shape was created by adapting a machine learning–generated Kirigami pattern from Ma et al.(2024) using simple geometric heuristics. The original pattern, shown in Fig.~\ref{fig:SI_MachineLearning}(a), was designed to produce a symmetric flower and consists of four identical regions labeled \circled{1}--\circled{4}. This pattern exhibits both $x$- and $y$-axis symmetry. In contrast, a butterfly shape is only symmetric about one axis. To achieve this, we reflected regions \circled{3} and \circled{4} about the $x$-axis to break the symmetry and introduce bilateral asymmetry (Fig.~\ref{fig:SI_MachineLearning}b). We then increased the aspect ratios of regions \circled{3} and \circled{4}, elongating them along the $y$-axis to mimic the larger hindwings of a butterfly (Fig.~\ref{fig:SI_MachineLearning}c). The final pattern in Fig.~\ref{fig:SI_MachineLearning}(c) was used to produce the butterfly shape presented in Fig.~4 of the main text.

